# Autonomous GIS: the next-generation AI-powered GIS

Zhenlong Li*, Huan Ning

[Geoinformation and Big Data Research Laboratory (GIBD)](#)
Department of Geography, University of South Carolina
709 Bull Street, Columbia, SC, USA 29208
*zhenlong@sc.edu

**Abstract**: Large Language Models (LLMs), such as ChatGPT, demonstrate a strong understanding of human natural language and have been explored and applied in various fields, including reasoning, creative writing, code generation, translation, and information retrieval. By adopting LLM as the reasoning core, we introduce Autonomous GIS as an AI-powered geographic information system (GIS) that leverages the LLM's general abilities in natural language understanding, reasoning, and coding for addressing spatial problems with automatic spatial data collection, analysis, and visualization. We envision that autonomous GIS will need to achieve five autonomous goals: self-generating, self-organizing, self-verifying, self-executing, and self-growing. We developed a prototype system called LLM-Geo using the GPT-4 API in a Python environment, demonstrating what an autonomous GIS looks like and how it delivers expected results without human intervention using three case studies. For all case studies, LLM-Geo was able to return accurate results, including aggregated numbers, graphs, and maps, significantly reducing manual operation time. Although still in its infancy and lacking several important modules such as logging and code testing, LLM-Geo demonstrates a potential path toward the next-generation AI-powered GIS. We advocate for the GIScience community to dedicate more effort to the research and development of autonomous GIS, making spatial analysis easier, faster, and more accessible to a broader audience.

## 1. Introduction

The rapid development of Artificial Intelligence (AI) has given rise to autonomous agents, also known as intelligent agents (Brustoloni 1991), which are computer systems capable of performing tasks and making decisions with minimal or no human intervention (Wooldridge and Jennings 1995). Autonomous agents have shown great potential in various domains, such as robotics, healthcare, transportation, and finance (Russell and Norvig, 2022). They offer numerous benefits, including increased efficiency, reduced errors, and the ability to handle complex tasks that would be time-consuming or impossible for humans to perform (Stone et al., 2022). Different from automatic systems that can perform expected tasks based on the given commands, inputs, and environmental settings (e.g., automatic doors and vending machines), autonomous agents (e.g., vacuum robots and autonomous vehicles) are typically more complex, more adaptable to environments and can make informed decisions based on the data it collects. The most challenging work for implementing autonomous agents is to build a decision-making core, which needs to be reacting appropriately according to its perception, while the perception and associated actions may not be pre-programmed. Most existing rule- or algorithm-based decision-making cores can only work in specified or closed environments, and their reasoning ability is limited.

The recent advancement of generative models, especially large language models (LLMs) such as GPT-3 (Brown et al. 2020) and GPT-4 (OpenAI 2023), has accelerated the research and development of autonomous agents. These models have demonstrated a strong understanding of human natural language, enabling them to perform tasks in various fields, including reasoning, creative writing, code generation, translation, and information retrieval. As a result, LLMs have been increasingly used as the decision-making core of digital autonomous agents, advancing the development of AI-powered systems. For example, Vemprala et al. (2023) used ChatGPT for robotics by testing various tasks, such as interaction, basic logical, geometrical, and mathematical reasoning. Researchers also used LLM to command pre-trained foundation models (Liang et al. 2023; Shen et al. 2023) to accomplish human tasks via natural languages, such as image manipulation (C. Wu et al. 2023) and captioning (Orlo 2023). Meanwhile, digital autonomous agents backed by LLMs have been developed, such as AutoGPT (Richards, 2023), BabyAGI (Nakajima, 2023), and AgentGPT (Reworkd, 2023). As an AI-freelancer platform, NexusGPT ("NexusGPT", 2023) released various digital autonomous agents to serve customers, providing services such as business consulting and software development.

The GIScience community has been incorporating AI into geospatial research and applications in recent years, leading to GeoAI, a subfield focusing on the intersection of AI and GIScience (VoPham et al. 2018). While GeoAI has made significant strides in utilizing AI for various geospatial applications such as spatial data processing and mining, the exploration and adoption of artificial general intelligence (AGI) (e.g., LLM as an early stage of AGI) in spatial analysis and GIS remains in its early stages. One notable by Mai et al. (2023) explores opportunities and challenges of developing an LLM-like foundation model for GeoAI. In a recent envision of GIS from the humanistic aspect, Zhao (2022) classified GIS into four categories: Embodiment GIS, Hermeneutic GIS, Autonomous GIS, and Background GIS by considering their relation (distance) to human, GIS, and place. Zhao regarded autonomous GIS "as either an independent agent or a place", such as drones, robot vacuums, and autonomous vehicles, or models trained to recognize land objects in images (e.g., buildings). Other attempts include using LLM to automate simple data operations (e.g., loading) (Kyriakou 2023; Mahmood 2023) in QGIS or adopting pre-trained models to segment images (Q. Wu 2023).

In this paper, we take a different approach to explore the integration of AI and GIS by narrowing down the connotation of Autonomous GIS as an AI-powered GIS that leverages LLM's general abilities in natural language understanding, reasoning, and coding for addressing spatial problems with automatic spatial data collection, analysis, and visualization. While autonomous GIS follows a similar concept to autonomous agents, most existing agents discussed above focus on text-based information retrieval, analysis, and summation, such as for business reports and travel planning, which do not necessarily have a unique, correct answer. Their design principles and architectures cannot be well adapted for GIS and spatial analysis, which are data-intensive and typically have only one correct answer, e.g., the population living within 10 kilometers of a hospital. We suggest that autonomous GIS should be designed as a programming- and data-centric framework to use automatic coding to address the GIScience questions of deterministic nature.

To demonstrate the feasibility of autonomous GIS, we developed a proof-of-concept prototype called LLM-Geo, which can conduct spatial analysis in an autonomous manner. LLM-Geo receives tasks (spatial problems/questions) from users and generates a solution graph (geoprocessing workflow) by decomposing the task into successive connected data operations as a directed acyclic graph. Each operation is a function to be implemented by the LLM, which is GPT-4 in this study. Next, LLM-Geo generates a combined program by integrating the codes of all operations and executes the combined program to produce the final result of the task. Three case studies are used to test the ability of LLM-Geo. The results indicate that the integration of LLMs into GIS has the potential to revolutionize the field by automating complex spatial analysis tasks and making GIS technology more accessible to individuals without GIS backgrounds. LLM-Geo, a prototypical autonomous GIS, serves as a potential path toward the next generation of AI-powered autonomous GIS.

The remainder of this paper is organized as follows: Section 2 elaborates on the concept and design considerations of autonomous GIS as an AI-powered autonomous system. Section 3 introduces the implementation of LLM-Geo as the prototype of autonomous GIS, followed by three case studies in Section 4. Sections 5 and 6 discuss what we have learned from LLM-Geo, its limitations, and potential future research directions. Section 7 concludes the paper.

## 2. Autonomous GIS as an AI-powered autonomous system

Autonomous systems are designed to make decisions and perform tasks without human intervention. They can adapt to changing conditions, learn from their environments, and make informed decisions based on the data they collect. By incorporating LLMs (or AGI) as the decision-making core for generating strategies and steps to solve spatial problems, autonomous GIS will be capable of searching and retrieving needed spatial data either from extensive existing online geospatial data catalogs or collecting new data from sensors, and then using existing spatial algorithms, models, or tools (or developing new ones) to process gathered data to generate the final results, e.g., maps, charts, or reports. LLM is the "brain" of autonomous GIS or the "head" if equipped with environmental sensors, while executable programs (e.g., Python) can be considered as its digital "hands". Similar to other autonomous agents (Sifakis 2019), we propose that autonomous GIS requires five critical modules, including decision-making (LLM as the core), data collecting, data operating, operation logging, and history retrieval. These modules enable GIS to achieve five autonomous goals: self-generating, self-organizing, self-verifying, self-executing, and self-growing (Table 1).

Unlike many autonomous agents for non-deterministic tasks with no strict standards to assess these answers such as writing a poem, GIScience applications require quantitative computation and analysis of spatial data to provide answers. Deterministic systems such as GIS are those in which the system's future state can be precisely predicted, given sufficient information about its initial conditions and the underlying rules governing its behavior. In a deterministic system, there is a direct relationship between the initial conditions and the outcome, with no room for randomness or uncertainty. In this sense, autonomous GIS needs a strictly controllable and explainable approach for the answer, and such answers should be unique and quantitatively correct based on the given data, e.g., the population living within 10 kilometers of a hospital. Therefore, we suggest that autonomous GIS should be designed as a programming-centric framework to use automatic coding to address the GIScience questions of deterministic nature.

Table 1 Autonomous goals and modules of autonomous GIS

| Autonomous Goal | Involving Modules | Functionality |
| --- | --- | --- |
| Self-generating | Decision-making, Data collecting | Generate solutions and data operation programs |
| Self-organizing | Decision-making, Operation logging | Ensure the operations are executed in the correct order, and data is stored in an appropriate manner |
| Self-verifying | Decision-making, Data operating | Test and verify the generated workflow, code, and programs |
| Self-executing | Data operating, Operation logging | Execute generated workflows, code, or programs |
| Self-growing | Operation logging, History retrieval | Reuse the verified operations |

## 3. LLM-Geo: a prototype of Autonomous GIS

We implemented two critical modules of autonomous GIS in LLM-Geo: decision-making and data operating, achieving three autonomous goals: self-generating, self-organizing, and self-executing. Additional modules are currently under development. The decision-making module adopts an LLM (GPT-4 in this study) as a core, or a "brain", to generate step-by-step solution workflow and develop associated codes of each step for addressing various spatial questions. The data operating module is a Python environment to execute the generated code, such as spatial data loading, processing, visualization, and saving.

Figure 1 shows the overall workflow of how LLM-Geo answers questions. The process begins with the user inputting the spatial question along with associated data locations such as online data URLs, REST (REpresentational State Transfer) services, and API (Application Program Interface) documentation. Then, the LLM generates a solution graph similar to a geoprocessing workflow. Based on the solution graph, LLM-Geo sends the requirements of each operation node to LLM, requesting code implementation. LLM-Geo then gathers all operation code implementations and asks LLM to generate an assembly program that connects the operations based on the workflow. Finally, LLM-Geo executes the assembly program to produce the final answer.

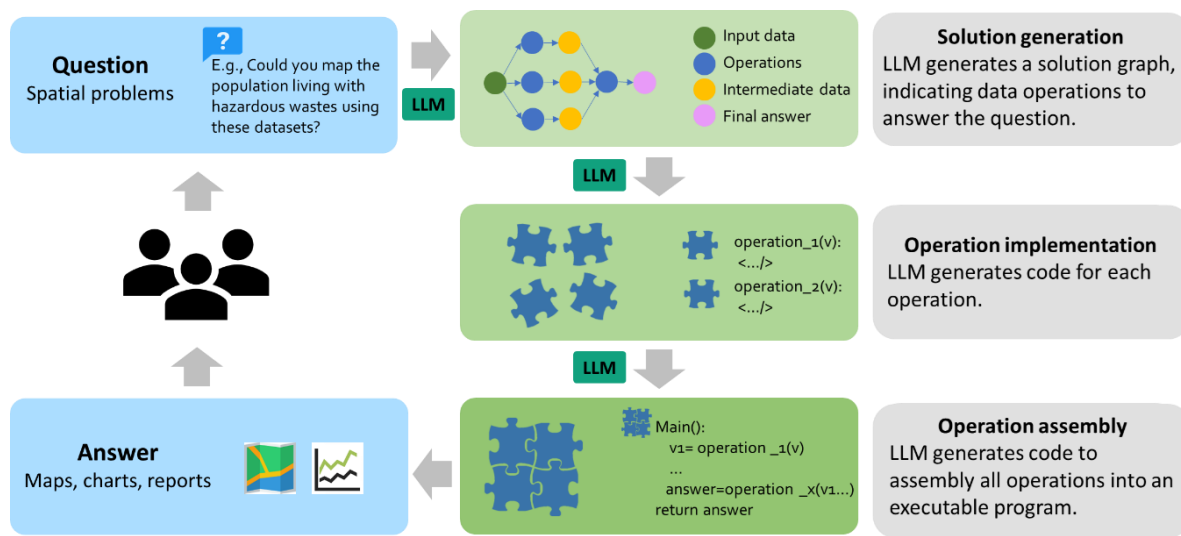

*Figure 1. The overall workflow of LLM-Geo*

### 3.1 Solution generation

A solution in LLM-Geo refers to a data processing workflow consisting of a series of connected operations and data. Performing these operations to process data in succession generates the final result for the question, such as maps, charts, tables, and new spatial datasets. We use a directed graph to represent the workflow and classify all nodes into two categories: data and operation. A *data node* refers to an operation's input data or output data, consisting of three types: input data node, intermediate data node, and output data node. The input data nodes tell the system where to load input data through local data paths, URLs, or REST APIs for programmatic data access. The output nodes are the final results for the task or question which can be numeric data, interactive or static maps, tables/charts, new datasets, or other user-requested types. All other data nodes are considered intermediate data nodes. An *operation node* is a process to manipulate data with input and output. Its inputs can be the input data nodes or intermediate data nodes from the ancestor operation nodes. Its output can be intermediate data nodes for descendent operation nodes or the output nodes (final result). The directed edges indicate the data flow. Disconnected nodes are not allowed in the solution graph, as all data flows will need to be converged at the output nodes to produce the final results. These structural constraints of the solution graph are fed to LLM (GPT-4) via API along with the question from the user. Since spatial analysis is essentially a geoprocessing workflow consisting of a series of connected spatial data processing tasks, the solution graph ensures that data flows seamlessly through the processing steps, ultimately converging at the final results (Figure 2). **Appendix 1** shows a sample of the prompt and LLM returned code for the solution graph.

Our experiments indicate that the granularity of an operation node is determined by LLM on-the-fly based on the complexity of the question and the maximum token length supported. Due to the token length limit, the LLM may have difficulty decomposing complex tasks into detailed graphs, resulting in a solution graph with lower granularity and fewer nodes. As all current LLMs have token limits (including state-of-the-art models such as GPT-4), this practical constraint necessitates a recursive approach for complex tasks. This approach can further decompose an operation node into a sub-solution graph containing more granular operations until the granularity reaches an appropriate level for accurate code generation and reusability.

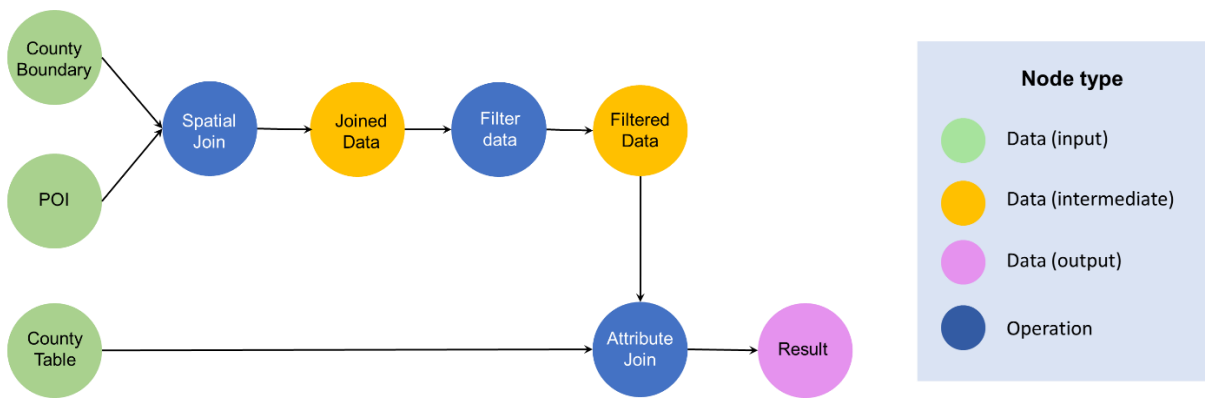

*Figure 2. Illustration of a solution graph with different node types*

## 3.2 Operation implementation

The generated solution graph serves as a "data processing plan" for a specified spatial question where the input data, output data, and data operations are pre-defined. In the operation implementation stage, each operation will be implemented as an executable code snippet following the operation definition in the solution graph. We use Python runtime as the data operation module in LLM-Geo, so the operation implementation is to generate Python code by LLM. LLM-Geo uses an algorithm to create interfaces between nodes to ensure the operation nodes and data nodes connect to their adjacent nodes. For example, the current version of LLM-Geo generates a Python function for each operation; the function definition and return data (i.e., names of functions and function input/output variables) are pre-defined in the solution graph. This strategy is used to reduce the uncertainty of code generation.

According to our experiments, GPT-4 requires sufficient information and extra guidance for reliable code generation. In this context, all information related to the solution graph paths preceding the current operation node may be needed. For example, an ancestor node might create a new column or file, and LLM must know and use the name of that column and file when generating function code for the current operation node. Additionally, information about descendant nodes is also needed to guide the code generation process. Therefore, LLM-Geo feeds the generated code in ancestor nodes and information of descendant nodes to LLM to request code implementation for each operation node. Besides node information, extra guidance is necessary. For instance, we found that GPT-4 has hazy memory regarding the prerequisites of some spatial data operations, such as the same column data type for table joining and identical map projection for overlay analysis. GPT-4 has learned to use GeoPandas, a popular Python library for spatial data processing; however, it seems unaware that this library does not support on-the-fly reprojection. In such cases, extra guidance is needed for LLM to generate correct code implementation for operation nodes. Table 2 presents some guidance that LLM-Geo used in the operation implementation stage. **Appendix 2** provides a sample of the prompt and LLM returned code for an operation.

*Table 2 Example guidance for operation code generation*

| | |
|---|---|
| 1 | DO NOT change the given variable names and paths. |
| 2 | Put your reply into a Python code block(enclosed by ```python and ```), NO explanation or conversation outside the code block. |
| 3 | If using GeoPandas to load zipped ESRI files from a URL, load the file directly, DO NOT unzip ESRI files. E.g., gpd.read_file(URL) |
| 4 | Generate descriptions for input and output arguments. |
| 5 | You need to receive the data from the functions, DO NOT load in the function if other functions have loaded the data and returned it in advance. |
| 6 | Note module 'pandas' has no attribute 'StringIO' |
| 7 | Use the latest Python module methods. |
| 8 | When doing spatial analysis, convert the involved layers into the same map projection. |
| 9 | When joining tables, convert the involved columns to string type without leading zeros. |
| 10 | When doing spatial joins, remove the duplicates in the results. Or please think about whether it needs to be removed. |

### 3. 3 Operation assembly

After generating the code (functions) for all operation nodes, LLM-Geo collects the codes and submits them to LLM, along with the solution graph and pre-defined guidance, to create a final program for the task. Based on the guidance, LLM uses the solution graph to determine the execution sequence of the operation nodes in the final program. Intermediate variables are generated to store the output data of operation nodes, which are then fed to the subsequent operation nodes. Finally, LLM-Geo executes the assembly program to produce the final result of the spatial question. The entire implementation structure and workflow of LLM-Geo based on GPT-4 API is shown in Figure 3. **Appendix 3** includes a sample of the prompt and LLM returned code for an assembly program.

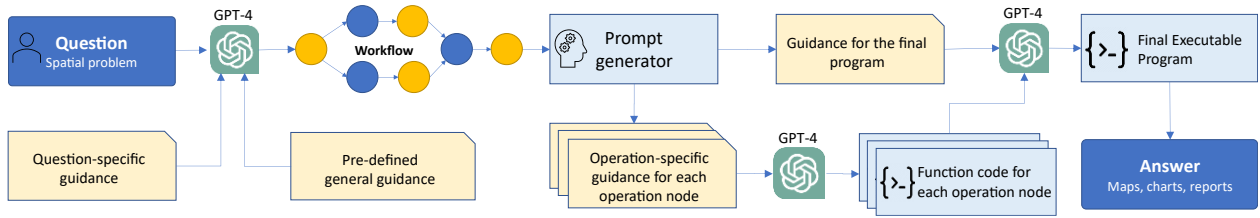

*Figure 3. Implementation workflow of LLM-Geo with GPT-4 API*

## 4. Case Studies

### 4.1 Case 1: Counting population living near hazardous wastes.

This spatial problem is to find out the population living with hazardous wastes and map their distribution. The study area is North Carolina, United States (US). We input the task (question) to LLM-Geo as shown in Box 1.

---

*Task:*
*1) Find out the total population that lives within a Census tract that contain hazardous waste facilities. The study area is North Carolina, US.*
*2) Generate a map to show the spatial distribution of population at the tract level and highlight the borders of tracts that have hazardous waste facilities.*

*Data locations:*
*1. NC hazardous waste facility ESRI shape file location: https://github.com/gladcolor/LLM-Geo/raw/master/overlay_analysis/Hazardous_Waste_Sites.zip*
*2. NC tract boundary shapefile location: https://github.com/gladcolor/LLM-Geo/raw/master/overlay_analysis/tract_shp_37.zip. The tract ID column is 'Tract'*
*3. NC tract population CSV file location: https://github.com/gladcolor/LLM-Geo/raw/master/overlay_analysis/NC_tract_population.csv. The population is stored in 'TotalPopulation' column. The tract ID column is 'GEOID'*

---

*Box 1. The spatial problem submitted to LLM-Geo for Case 1.*

This question asks LLM-Geo to find out the total population living with hazardous wastes and generate a map for the population distribution with sufficient details for the analysis, including the data location and used column names. Note that the vector layers (hazardous facilities and tract boundaries) are not in the same map projection, and the tract ID data type is *text* in the boundary layer, but it is *integer* when Pandas reads the population CSV file. GPT-4 has hazy memory to pre-process these inconsistencies, so LLM-Geo needs to remind GPT-4 with extra guidance in the prompt: "When doing spatial analysis, convert the involved layers into the same map projection. When joining tables, convert the involved columns to string type without leading zeros."

Figure 4 shows the outputs of LLM-Geo for this case study, including a solution graph detailing the data processing steps, the final assembly Python program, and the final results for the question: the total population (5,688,769) that lives within a tract that contains hazardous waste facilities and a map showing the spatial distribution of population at the Census tract level and highlight the borders of tracts that have hazardous waste facilities. Our manual verification confirmed the accuracy of the number and the map. With the working code generated from LLM-Geo, users can easily adjust and re-run the code to customize the map visualization styles if needed.

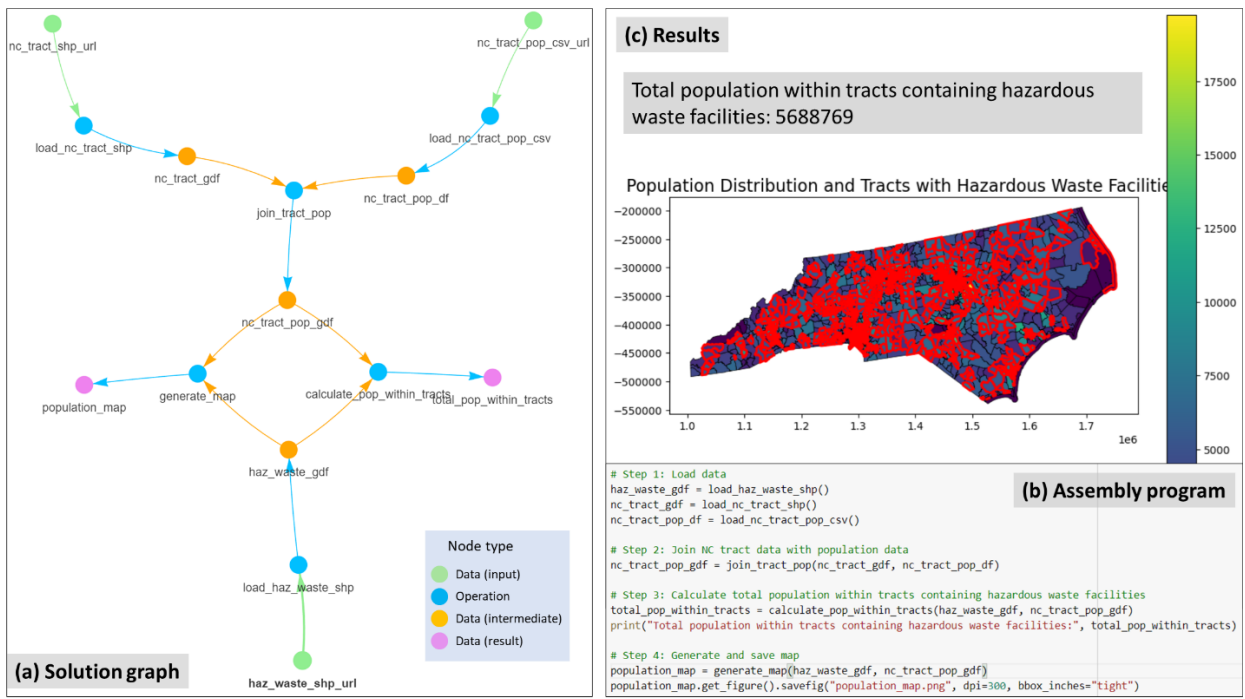

*Figure 4. Results automatically generated by LLM-Geo for counting the population living near hazardous wastes. (a) Solution graph, (b) assembly program (Python codes), and (c) returned population count and generated map.*

### 4.2 Case 2: Human mobility data retrieval and trend visualization

This task investigates the mobility changes during COVID-19 pandemic in France in 2020. First, we asked LLM-Geo to retrieve mobility data from the ODT Explorer using REST API (Li et al. 2021), and then compute and visualize the monthly change rate compared to January 2020. We input the task to LLM-Geo as shown in Box 2. Figure 5 shows the results from LLM-Geo for this case study, including a solution graph detailing the data processing steps, a map matrix showing the spatial distribution of the mobility change rate, a line chart showing the trend of the mobility change rate, and the final assembly program that produced the outputs. All results are correct per our manual verification.

*Task:*
*1) Show the monthly change rates of population mobility for each administrative regions in a France map. Each month is a sub-map in a map matrix. The base of the change rate is January 2020.*
*2) Draw a line chart to show the monthly change rate trends of all administrative regions. Th x-axis is month.*

*Data locations:*
*1. ESRI shapefile for France administrative regions: https://github.com/gladcolor/LLM-Geo/raw/master/REST_API/France.zip. The 'GID_1' column is the administrative region code, 'NAME_1' column is the administrative region name.*
*2. REST API URL with parameters for mobility data access: http://gis.cas.sc.edu/GeoAnalytics/REST?operation=get_daily_movement_for_all_places&source=twitter&scale=world_first_level_admin&begin=01/01/2020&end=12/31/2020. The response is in CSV format. There are three columns in the response: place, date (format:2020-01-07), and intra_movement. 'place' column is the administrative region code, France administrative regions start with 'FRA'.*

*Box 2. The spatial problem submitted to LLM-Geo for Case 2.*

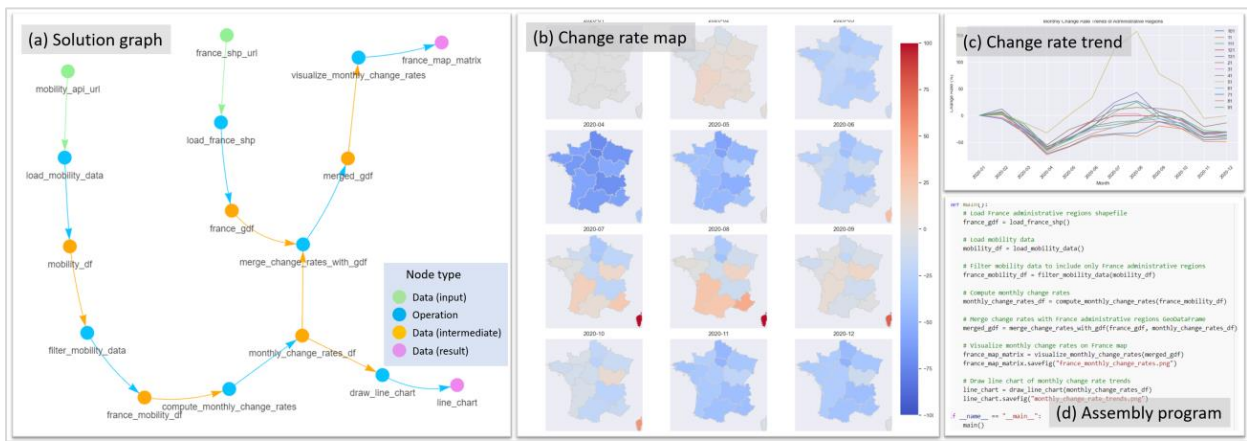

*Figure 5. Results automatically generated by LLM-Geo for human mobility data retrieval and trend visualization. (a) Solution graph, (b) map matrix showing the spatial distribution of mobility change rate, (c) line chart showing the trend of the mobility change rate, (d) assembly program.*

### 4.3 Case 3: COVID-19 death rate analysis and visualization at the US county level

The spatial problem for this case is to investigate the spatial distribution of the COVID-19 death rate (ratio of COVID-19 deaths to cases) and the association between the death rate and the proportion of senior residents (age >=65) at the US county level. The death rate is derived from the accumulated COVID-19 data as of December 31, 2020, available from New York Times (2023), based on state and local health agency reports. The population data is extracted from the 2020 ACS five-year estimates (US Census Bureau 2022). The task asks for a map to show the county level death rate distribution and a scatter plot to show the correlation and trend line of the death rate with the senior resident rate. Box 3 shows the spatial problem along with the needed dataset locations submitted to LLM-Geo. The results are presented in Figure 6.

*Task:*
*1) Draw a map to show the death rate (death/case) of COVID-19 among the countiguous US counties. Use the accumulated COVID-19 data of 2020.12.31 to compute the death rate. Use scheme ='quantiles' when plotting the map. Set map projection to 'Conus Albers'. Set map size to 15*10 inches.*
*2) Draw a scatter plot to show the correlation and trend line of the death rate with the senior resident rate, including the r-square and p-value. Set data point transparency to 50%, regression line as red. Set figure size to 15*10 inches.*

*Data locations:*
*1) COVID-19 data case in 2020 (county-level): https://github.com/nytimes/covid-19-data/raw/master/us-counties-2020.csv. This data is for daily accumulated COVID cases and deaths for each county in the US. There are 5 columns: date (format: 2021-02-01), county, state, fips, cases, deaths.*
*2) Contiguous US county boundary (ESRI shapefile): https://github.com/gladcolor/spatial_data/raw/master/contiguous_counties.zip. The county FIPS column is 'GEOID'.*
*3) Census data (ACS2020): https://raw.githubusercontent.com/gladcolor/spatial_data/master/Demography/ACS2020_5year_county.csv. The needed columns are: 'FIPS', 'Total Population', 'Total Population: 65 to 74 Years', 'Total Population: 75 to 84 Years', 'Total Population: 85 Years and Over'.*

*Box 3. The spatial problem submitted to LLM-Geo for Case 3.*

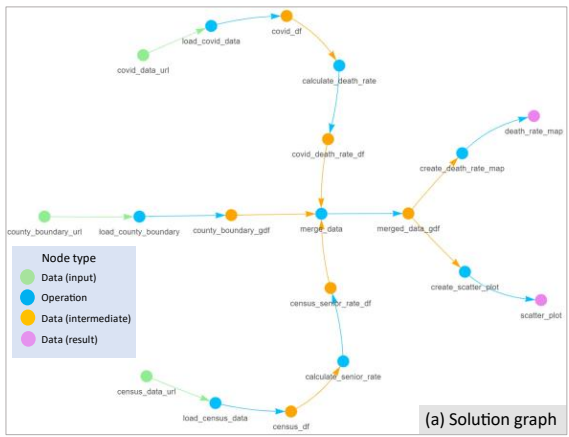

(a) Solution graph

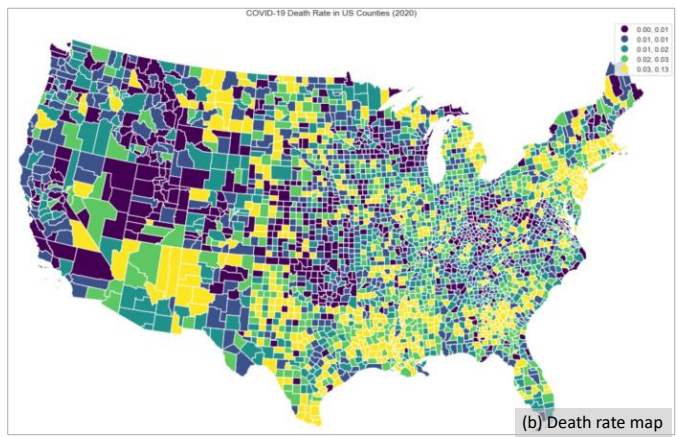

(b) Death rate map

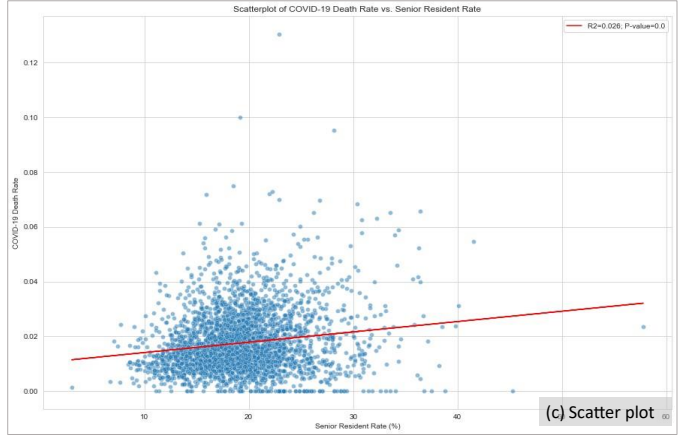

(d) Assembly program

(c) Scatter plot

*Figure 6. Results automatically generated by LLM-Geo for the US county level COVID-19 death rate analysis and visualization. (a) Solution graph, (b) county level death rate map of the contiguous US, (c) scatter plot showing the association between COVID-19 death rate and the senior resident rate at the county level, (d) assembly program.*

In this example, we provided more requirements on the visualization style of the map and chart, such as *"Use scheme ='quantiles' when plotting the map. Set map projection to 'Conus Albers'. Set map size to 15*10 inches".* Without such guidance, the system often picks different settings for each request of the same problem. The entire generated executable program can be found in **Appendix 4**. By changing the date "2020-12-31" in the code and re-run the program, one can easily explore the COVID-19 death rate distribution and its association with the senior resident rate for other periods.

## 5. Discussion and lessons learned

### 5.1 Forms of autonomous GIS

Autonomous GIS aims to mimic human operations in spatial analysis rather than changing its foundation and procedure. In this sense, autonomous GIS can function as a standalone application (local or cloud-based), acting as a GIS analyst who takes questions from users and produces answers. One potential graphic user interface (GUI) for such an application is illustrated in Figure 7 (a), where users input the description of a spatial problem, click the *Submit* button, and receive the results directly within the application. The application should also be able to display the generated solution graph (geoprocessing workflow) and executable codes that produce the final results for monitoring, debugging, and customization purposes. Autonomous GIS's ability can self-grow rapidly by accumulating and reusing the generated code and workflows, especially the cloud version that serves many users.

In addition to the standalone form, autonomous GIS can serve as a co-pilot for traditional GIS software, using natural language to communicate with users and automate spatial data processing and analysis tasks. This setup is similar to Microsoft's Copilot for its Office family, which automates office tasks such as report writing and slide creation (Microsoft 2023). For example, an autonomous GIS panel alongside the map view can be integrated into ArcGIS and QGIS, displaying the chatbox, generated solution graph, and codes as illustrated in Figure 7 (b). The result will be shown in the built-in map view. Users can modify the workflow by editing the solution graph or operation parameters. If an operation is implemented by generated code, users can locate and edit the code by clicking the solution graph nodes. Overall, the solution graph is similar to the Module Builder of ArcGIS or Model Designer in QGIS. Implementing autonomous GIS based on existing GIS platforms may be the most practical and efficient approach at present, as mature GIS platforms (e.g., ArcGIS) already have a rich number of operations (e.g., the tools in the ArcGIS toolbox) along with well-established documentation that LLM can quickly learn and use to generate the solution graph. More broadly, autonomous GIS can be used as a "plugin" integrated with applications requiring geospatial data manipulation, such as a public health surveillance dashboard for COVID-19 or other diseases. Such an autonomous GIS-enabled dashboard will have the ability to interact with users using natural language to generate customized maps, charts, reports, new datasets based on the users' specific needs and preferences. This level of interaction and customization allows users to have a more personalized and efficient experience when working with geospatial data in various applications.

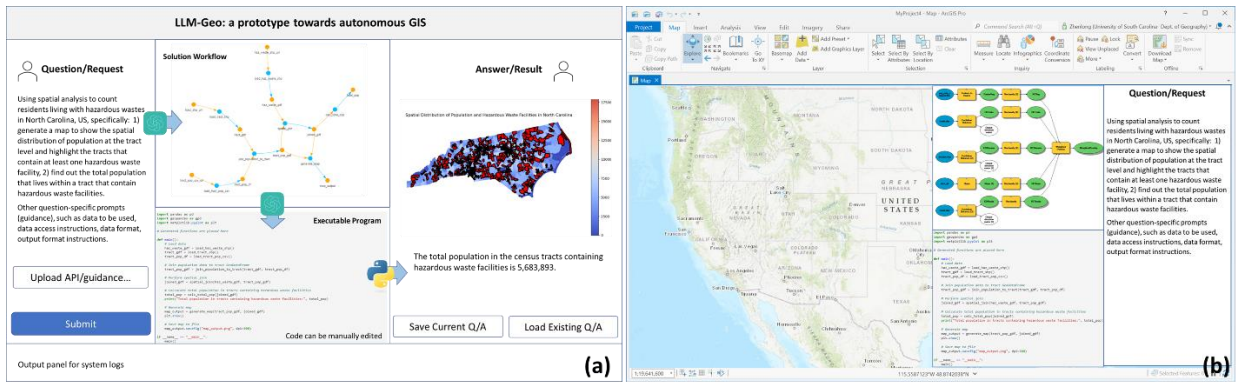

*Figure 7. Illustration of the user interface for (a) standalone autonomous GIS, (b) co-pilot style autonomous GIS integrated in existing GIS software (e.g., ArcGIS Pro). Note: the question, workflow, codes and results in this figure are made up for demonstration purposes only.*

**5.2 Difference between autonomous GIS and AI assistants**

There are various AI assistants backed by LLMs available in the market. For example, Cursor (Cursor 2023) and GitHub Copilot X (GitHub 2023) are similar products that utilize LLMs to generate code and comments to enhance programmers' productivity. Microsoft's Copilot for its Office family is able to automate office tasks such as report writing and slide creation. What distinguishes autonomous GIS from these AI assistants lies in the inputs, final products, and the deterministic and data-centric workflow generating the results. The AI assistants often receive text and generate text-based content for users, such as paragraphs or code. In contrast, the input for GIS analysis is spatial data, which is further processed to derive quantitative results (e.g., maps, charts) to support decision-making.

Furthermore, GIS users typically create geoprocessing workflows (Scheider et al. 2021) involving spatial data and GIS tools based on their analysis objectives, experience, and knowledge. These workflows can take the form of scripts, graphically connected data and operations using a model builder, or a sequence of manual steps in GIS software. These workflows are deterministic, data-centric, and can be replicated by others. The design of the autonomous GIS prototype (LLM-Geo) follows a similar problem-solving approach by first generating a data-centric geoprocessing workflow, implementing the workflow with programming, and executing the workflow to produce the final results in an automated manner. It is important to note that LLM does not directly manipulate spatial data; instead, it generates and commands operations (programs) to handle the data.

**5.3 Divide-and-conquer**

One might reasonably question why we do not opt for a more direct approach — asking LLMs like GPT-4 for solutions without generating a solution graph and operations. Indeed, our tests have shown that with extra guidance embedded in the prompt, GPT-4 is capable of generating correct code for relatively simple spatial analysis tasks. However, this direct method may limit the LLM's capacity to tackle more complex tasks such as assessing the accessibility of Autism intervention services at a national level using smartphone mobility data APIs and web crawling or identifying the most promising fishing site in a specific sea area. These tasks are akin to writing a long novel, and it is unlikely that LLMs would generate comprehensive, one-step solutions that involve extensive programs. Such tasks are challenging not only for humans but also for models (Maeda, 2023).

Human problem-solving often employs a divide-and-conquer strategy when facing complex tasks. By adopting this approach in the design of LLM-Geo, it is intended to address spatial analysis tasks by breaking complex problems into smaller, more manageable sub-problems that LLMs can handle (i.e., operations). It systematically addresses these sub-problems (i.e., developing a function for each sub-problem) and ultimately combines these functions (i.e., assembling a program) to yield the final results. This divide-and-conquer methodology has been employed in other digital autonomous agents, such as AutoGPT (Richards 2023) and AgentGPT (Reworkd 2023). We plan to further explore the feasibility of this approach with tasks that are more complex than those demonstrated thus far. Additionally, the divide-and-conquer strategy aids in the production of verified operation nodes (e.g., code snippets). These validated sub-solutions (operations) become valuable assets for autonomous GIS as they can be repurposed for future tasks, bolstering the autonomous goal of self-growing. Furthermore, this approach facilitates error location in operation nodes, which eases human intervention when necessary.

**5.4 The need for sufficient information in GIS and LLM**

As previously discussed, GIS is deterministic in nature that allows for precise predictions of their future states given enough knowledge about their starting conditions and the rules that govern their operations. In these deterministic systems, outcomes are intrinsically tied to their initial conditions, leaving no room for randomness or uncertainty. Sufficient information is vital in deterministic systems, as it facilitates accurate prediction and comprehension of the system's behavior. For GIS, spatial analysis usually faces constraints such as data availability, jurisdiction boundary, and map size. Furthermore, data metadata or details like layer projections, field names, and data types, are critical for subsequent spatial programming. Much like a human analyst, autonomous GIS needs to be aware of these constraints and information to successfully accomplish the given task. For example, one is unable to perform an attribute join operation without prior knowledge of the column names or conduct an overlay analysis without information on layer projections. Therefore, we believe that the provision of sufficient information is both necessary and sufficient condition for autonomous GIS's accurate reasoning, task planning, and action execution. Possible methods to supply such information include data sampling, web searches, self-reflection, and manual provision.

**5.5 Recall the hazy memory of LLM**

LLMs can correct previous error output by incorporating additional information. They understand the steps required to complete a task, but often overlook practical constraints in spatial analysis, such as the necessity of matching map projection for overlay analysis and precise data type matching for the common joining column of two attribute tables. This sort of "hazy" memory in LLM likely results from limited GIS training materials. To enhance LLMs' recall of these hazy memories, and thus improving their capacity for successful reasoning and coding, it is crucial to embed necessary reminders within the task description. Despite the potential variation in this hazy spatial analysis memory across different LLMs and training datasets, it is feasible to formulate a relatively universal reminder list to aid autonomous GIS in producing accurate results. This list functions similar to a checklist for GIS users. For instance, during spatial analysis tasks, GIS users must ensure that the common joining columns have the same data type (string or int) and leading zero digits. Experienced GIS users typically verify these prerequisites in advance, while novices may overlook these steps until they encounter errors or consult a checklist. The specificity of this guidance, or checklist, will vary among users (e.g., shorter for experienced GIS users) and will also differ for autonomous GIS depending on the sophistication of the employed LLMs.

# 6. Limitations and future work

While LLM-Geo validates the concept of autonomous GIS, it is still in its infancy with a number of limitations, such as the inability to debug the generated or to process raster data. The advent of reasoning capabilities in LLMs brings the GIScience community to the dawn of a new era, yet there is still a vast landscape of opportunities to explore and challenges to address. We have identified several potential avenues for further research and development to tackle the existing constraints of LLM-Geo and lay the groundwork for a fully operational autonomous GIS in the future.

## 6.1 The adaptivity of LLM-Geo needs to be improved

We have developed a number of case studies to test LLM-Geo and reported three typical cases in Section 4 with a success rate of about 80% based on GPT-4. Our observations indicate that the generated solution graphs were typically accurate. The majority of failures could be traced back to faulty code during the operation implementation stage; even with the divide-and-conquer approach, a single error statement can crash the entire program. Although GPT-4 usually utilizes the wide-used GeoPandas library (Jordahl et al. 2020) for spatial analysis, it often struggles with generating correct code in a single attempt. Furthermore, the rate of errors increases when it uses other libraries such as Plotly (2023). Currently, LLM-Geo lacks a mechanism to review, test, debug, and verify the generated code. These modules are needed to achieve the autonomous goal of *self-verifying*. A preliminary implementation may be directly pushing code and debugging errors to LLMs to request corrected code.

Another observation is that LLM-Geo tends to develop complex functions from scratch based on GeoPandas or other open-source spatial libraries for a spatial operation (e.g., nearest distance calculation) that is already available in the traditional and well-developed GIS packages or toolboxes (e.g., QGIS and ArcGIS). Such on-the-fly generated complex functions are error-prone, even with appropriate guidance. Compared to the current code generation approach, we believe that LLM-Geo could be more adaptive and robust by utilizing the well-established and tested GIS toolboxes coupled with the code debugging and verification module. Another limitation is that LLM-Geo is currently incapable of knowing the data, such as map projections, data types, and attributes. A sophisticated mechanism needs to be designed to manage these interactions between LLM and code and data in an automated manner.

## 6.2 A memory system is needed

Every autonomous system needs a memory component to store the contextual and long-term information, data, and results for future retrieval or reuse, a feature that LLM-Geo currently lacks. While LLM-Geo does use a simple method to store contextual memory, including prompts and LLM responses, this is insufficient for an autonomous system. Autonomous GIS's memory can be data sources, pre-defined analysis guidance, and verified codes generated in previous tasks. These memory elements are expected to expand over time to achieve the self-growing autonomous goal. Most LLM-based autonomous agents (e.g., AutoGPT) use vector databases such as Pinecone (Pinecone 2023) to store and retrieve conversations. Autonomous GIS faces a more complicated requirement that more than text storage and retrieval it needs to store the solution graph (geoprocessing workflow), which would become hierarchical to recursively solve complex tasks. Moreover, the verified solutions and operation codes would become part of the long-term information to be memorized. Thus, a memory system along with a logging module is needed to achieve the autonomous goal of *self-growing*.

### 6.3 Categorized guidance maintained by GIS community

Despite being aware of the hazy memory exhibited by GPT-4, we have managed to supplement it with extra guidance, steering it toward generating accurate operation nodes for data manipulation. Like the documentation of open-sourced packages like GDAL/OGR, GIS guidance for LLMs contains the experience of GIS analysts and serves as a handbook for LLMs, which can play a critical role in automating spatial analysis. The current guidance is being updated throughout the development of LLM-Geo. However, such guidance summarized from a few cases is far from covering the most common spatial analysis scenarios. For example, the case studies did not include raster data analyses, which are typical in GIS applications. We invite the GIS community to contribute GIS guidance for LLMs to enhance their spatial programming (code generation) capabilities. Given the potential length of such guidance documents, they should be categorized. We could, for example, have separate guidance documents for the analysis of vector data, raster data, network data, and geo-visualization. The memory system of autonomous GIS should be designed to retrieve necessary guidance only for specific tasks rather than read them all. Researchers can also investigate how to summarize spatial analysis guidance based on the autonomous GIS's trials and errors, for example, exploring how to guide LLM to generate and execute various analyzing tasks to produce guidance using reinforcement learning.

### 6.4 Trial-and-error, a more robust problem-solving approach

The current implementation of LLM-Geo assumes that once all necessary information is fed into LLM, the correct result will be returned. This expectation is somewhat idealistic as spatial analysis is often complex in terms of data sources, algorithms, and hypotheses. In practice, analysts are more likely to adopt a trial-and-error problem-solving strategy, exploring various possible paths and pruning unsuccessful branches in the solution graph. This strategy mirrors the human problem-solving approach, as it is common for even experienced programmers to encounter errors during code development. Yao et al. (2023)'s experiments about "tree of thoughts" demonstrated substantial improvement in the reasoning of LLMs on the game of 24, creative writing, and mini crosswords. Their "tree of thoughts" is similar to a trial-and-error strategy: evaluate a possible branch, and then determine whether to cut it off. However, the simplicity of the tasks used in their experiments limits the generalizability of their findings in more complex domains such as spatial analysis.

### 6.5 Online geospatial data discover and filtering

Our case studies provided data for LLM-Geo, such as data file paths, URLs, or data access APIs, as well as data descriptions. LLM-Geo is ready to accept spatial analysis tasks with similar data input. However, an autonomous GIS should be capable of collecting required data independently to finish the task. Most LLM-based autonomous agents are equipped with search engines (using APIs) to search and retrieve data from the internet. Fortunately, numerous geospatial datasets, standard geospatial web services (e.g., OGC WMS, WFS, WCS, WPS), and REST APIs for geospatial data access have been established in recent decades. For example, US population data can be extracted from the US Census Bureau (2023) via API, as well as from OpenStreetMap (OpenStreetMap 2023). Large online geospatial data catalogs have also been created by various national or international organizations and government agencies, such as the Google Earth Engine Data Catalog, NOAA and NASA Data Catalogs, EarthCube, and the European Environment Agency (EEA) geospatial data catalog. The challenge lies in finding and selecting suitable and high-quality data layers in terms of spatiotemporal resolution, spatiotemporal coverage, and accuracy. Autonomous GIS developers need to establish practical strategies for LLMs to discover, filter, and utilize the most relevant and accurate geospatial datasets for a given spatial analysis task.

### 6.6 Answer "why" questions

LLM-Geo, powered by GPT-4, illustrates the ability of knowing 'how' to perform spatial analysis and automate a multitude of routine geospatial tasks. The next challenge lies in the ability of autonomous GIS to address 'why' questions, which often requires a more profound understanding and investigation of geospatial knowledge. Questions such as "According to the smartphone mobility data, why did my customer numbers drop by 20% this month?" or "Why did these migratory birds alter their path in the past decade?" extend beyond basic spatial analysis and delve into the domain of hypothesis generation, data selection, and experimental design. To answer these questions, autonomous GIS needs to have the ability to design research by formulating informed hypotheses based on the posed question, available data, and context.

### 6.7 Build a Large Spatial Model (LSM)

LLMs, trained on extensive text corpora, have developed language skills, knowledge, and reasoning abilities, but their spatial awareness remains limited due to the scarcity of spatial samples within these corpora. Many resources in GIScience, such as abundant historical remote sensing images, global vector data, detailed records of infrastructure and properties, and vast amounts of other geospatial big data sources, have yet to be fully incorporated into the training of large models. Consider the potential of a Large Spatial Model (LSM), trained on all available spatial data, mirroring the way LLMs are trained on extensive text corpora. Such a model could potentially possess a detailed understanding of the Earth's surface, accurately describe any location, and comprehend the dynamics of ecosystems and geospheres. This would not only enrich the spatial awareness of future artificial general intelligence, but could also transform the field of GIScience, empowering autonomous GIS to answer "why" questions. We advocate for further research and efforts towards the training of large spatial models that can more accurately represent the Earth's surface and human society.

## 7. Conclusion

In this study, we presented Autonomous GIS as the next-generation of AI-driven geographic information systems, aimed at making GIS and spatial analysis more accessible and user-friendly. The goal of autonomous GIS is to accept tasks via natural language and solve spatial problems with minimal to no human intervention. By employing LLM as the core reasoning mechanism, we suggest that autonomous GIS should achieve five autonomous objectives: self-generating, self-organizing, self-verifying, self-executing, and self-growing. To illustrate this concept, we developed a prototype of autonomous GIS, LLM-Geo, using the GPT-4 API within a Python environment, demonstrating what an autonomous GIS could look like and how it can deliver results autonomously. In three case studies, LLM-Geo successfully produced the expected results, such as aggregated numbers, charts, and maps, significantly reducing manual operation time. Although still in its infancy, LLM-Geo demonstrates the concept and feasibility of autonomous GIS. Echoing the vision of Zhu et al. (2021) that the "Next generation of GIS must be easy", we believe autonomous GIS as the next-generation AI-powered GIS holds great promise towards achieving this goal. We encourage the GIScience community to dedicate more effort to the research and development of autonomous GIS, making spatial analysis easier, faster, and more accessible to a broader audience.

**Data Availability Statement:** The source code of LLM-Geo and case study data are provided at
https://github.com/gladcolor/LLM-Geo

**Acknowledgements:** We thank Professor Gregory J. Carbone for helping revise an earlier draft of this manuscript.

# Appendix
## Sample prompts and responses of LLM-Geo

### 1. Solution graph generation

Prompt (Note: The spatial problem indicated in the blue text is the only input needed from users, other text is pre-defined in LLM-Geo)

> Your role: A professional Geo-information scientist and developer good at Python.
> Task: Generate a graph (data structure) only, whose nodes are (1) a series of consecutive steps and (2) data to solve this question:
> 1) Find out the total population that lives within a tract that contain hazardous waste facilities. The study area is North Carolina, US.
> 2) Generate a map to show the spatial distribution of population at the tract level and highlight the borders of tracts that have hazardous waste facilities.
> Data locations (each data is a node):
> 1. NC hazardous waste facility ESRI shape file location: https://github.com/gladcolor/LLM-Geo/raw/master/overlay_analysis/Hazardous_Waste_Sites.zip.
> 2. NC tract boundary shapefile location: https://github.com/gladcolor/LLM-Geo/raw/master/overlay_analysis/tract_shp_37.zip. The tract id column is 'Tract'.
> 3. NC tract population CSV file location: https://github.com/gladcolor/LLM-Geo/raw/master/overlay_analysis/NC_tract_population.csv. The population is stored in 'TotalPopulation' column. The tract ID column is 'GEOID'.
>
> Your reply needs to meet these requirements:
> 1. Think step by step.
> 2. Steps and data (both input and output) form a graph stored in NetworkX. Disconnected components are NOT allowed.
> 3. Each step is a data process operation: the input can be data paths or variables, and the output can be data paths or variables.
> 4. There are two types of nodes: a) operation node, and b) data node (both input and output data). These nodes are also input nodes for the next operation node.
> 5. The input of each operation is the output of the previous operations, except the those need to load data from a path or need to collect data.
> 6. You need to carefully name the output data node.
> 7. The data and operation form a graph.
> 8. The first operations are data loading or collection, and the output of the last operation is the final answer to the task. Operation nodes need to connect via output data nodes, DO NOT connect the operation node directly.
> 9. The node attributes include: 1) node_type (data or operation), 2) data_path (data node only, set to "" if not given), and description. E.g., {'name': "County boundary", "data_type": "data", "data_path":"D:\Test\county.shp",  "description": "County boundary for the study area"}.
> 10. The connection between a node and an operation node is an edge.
> 11. Add all nodes and edges, including node attributes to a NetworkX instance, DO NOT change the attribute names.
> 12. DO NOT generate code to implement the steps.
> 13. Join the attribute to the vector layer via a common attribute if necessary.

14. Put your reply into a Python code block, NO explanation or conversation outside the code block (enclosed by ```python and ```).
15. Note that GraphML writer does not support class dict or list as data values.
16. You need spatial data (e.g., vector or raster) to make a map.
17. Do not put the GraphML writing process as a step in the graph.
18. Save the network into GraphML format, save it at: E:\Research\LLM-Geo\Resident_at_risk_counting\Resident_at_risk_counting.graphml

Reply example:
```python
import networkx as nx
G = nx.DiGraph()
# Add nodes and edges for the graph
# 1 Load hazardous waste site shapefile
G.add_node("haz_waste_shp_url", node_type="data", path="https://github.com/gladcolor/LLM-Geo/raw/master/overlay_analysis/Hazardous_Waste_Sites.zip", description="Hazardous waste facility shapefile URL")
G.add_node("load_haz_waste_shp", node_type="operation", description="Load hazardous waste facility shapefile")
G.add_edge("haz_waste_shp_url", "load_haz_waste_shp")
G.add_node("haz_waste_gdf", node_type="data", description="Hazardous waste facility GeoDataFrame")
G.add_edge("load_haz_waste_shp", "haz_waste_gdf")
...
```

**Generated Solution Graph Code:**

```python
import networkx as nx
G = nx.DiGraph()
# 1. Load hazardous waste site shapefile
G.add_node("haz_waste_shp_url", node_type="data", data_path="https://github.com/gladcolor/LLM-Geo/raw/master/overlay_analysis/Hazardous_Waste_Sites.zip", description="Hazardous waste facility shapefile URL")
G.add_node("load_haz_waste_shp", node_type="operation", description="Load hazardous waste facility shapefile")
G.add_edge("haz_waste_shp_url", "load_haz_waste_shp")
G.add_node("haz_waste_gdf", node_type="data", description="Hazardous waste facility GeoDataFrame")
G.add_edge("load_haz_waste_shp", "haz_waste_gdf")
# 2. Load NC tract boundary shapefile
G.add_node("nc_tract_shp_url", node_type="data", data_path="https://github.com/gladcolor/LLM-Geo/raw/master/overlay_analysis/tract_shp_37.zip", description="NC tract boundary shapefile URL")
G.add_node("load_nc_tract_shp", node_type="operation", description="Load NC tract boundary shapefile")
G.add_edge("nc_tract_shp_url", "load_nc_tract_shp")
G.add_node("nc_tract_gdf", node_type='data', description="NC tract boundary GeoDataFrame")
G.add_edge("load_nc_tract_shp", "nc_tract_gdf")
# 3. Load NC tract population CSV file
G.add_node("nc_tract_pop_csv_url", node_type="data", data_path="https://github.com/gladcolor/LLM-Geo/raw/master/overlay_analysis/NC_tract_population.csv", description="NC tract population CSV file URL")
G.add_node("load_nc_tract_pop_csv", node_type="operation", description="Load NC tract population CSV file")
G.add_edge("nc_tract_pop_csv_url", "load_nc_tract_pop_csv")
G.add_node("nc_tract_pop_df", node_type="data", description="NC tract population DataFrame")
G.add_edge("load_nc_tract_pop_csv", "nc_tract_pop_df")
# 4. Join tract GeoDataFrame with population DataFrame
G.add_node("join_tract_pop", node_type="operation", description="Join tract GeoDataFrame with population DataFrame")
G.add_edge("nc_tract_pop_df", "join_tract_pop")
G.add_edge("nc_tract_gdf", "join_tract_pop")
G.add_node("nc_tract_pop_gdf", node_type="data", description="NC tract GeoDataFrame with population")
```

```python
G.add_edge("join_tract_pop", "nc_tract_pop_gdf")
# 5. Identify and calculate the total population within tracts containing hazardous waste
facilities
G.add_node("calculate_pop_within_tracts", node_type="operation", description="Calculate total
population within tracts containing hazardous waste facilities")
G.add_edge("nc_tract_pop_gdf", "calculate_pop_within_tracts")
G.add_edge("haz_waste_gdf", "calculate_pop_within_tracts")
G.add_node("total_pop_within_tracts", node_type="data", description="Total population within
tracts containing hazardous waste facilities")
G.add_edge("calculate_pop_within_tracts", "total_pop_within_tracts")
# 6. Generate the map showing spatial distribution of population and highlighting borders of
tracts with hazardous waste facilities
G.add_node("generate_map", node_type="operation", description="Generate the map showing spatial
distribution of population and highlighting borders of tracts with hazardous waste facilities")
G.add_edge("nc_tract_pop_gdf", "generate_map")
G.add_edge("haz_waste_gdf", "generate_map")
G.add_node("population_map", node_type="data", description="Spatial distribution of population
and highlighted borders of tracts with hazardous waste facilities")
G.add_edge("generate_map", "population_map")
# Save the graph
nx.write_graphml(G, "E:\\Research\\LLM-
Geo\\Resident_at_risk_counting\\Resident_at_risk_counting.graphml")
```

## 2. Operation code generation

Prompt to generate the function for one operation (note that this prompt is automatically generated by LLM-Geo based on the question and pre-defined guidance):

> Your role: A professional Geo-information scientist and developer good at Python.
> operation_task: You need to generate a Python function to do: Join tract GeoDataFrame with population DataFrame
> This function is one step to solve the question: 1) Find out the total population that lives within a tract that contains hazardous waste facilities. The study area is North Carolina, US.
> 2) Generate a map to show the spatial distribution of population at the tract level and highlight the borders of tracts that have hazardous waste facilities.
> Data locations: 1. NC hazardous waste facility ESRI shape file location: https://github.com/gladcolor/LLM- Geo/raw/master/overlay_analysis/Hazardous_Waste_Sites.zip.
> 2. NC tract boundary shapefile location: https://github.com/gladcolor/LLM-Geo/raw/master/overlay_analysis/tract_shp_37.zip. The tract id column is 'Tract'.
> 3. NC tract population CSV file location: https://github.com/gladcolor/LLM-Geo/raw/master/overlay_analysis/NC_tract_population.csv. The population is stored in 'TotalPopulation' column. The tract ID column is 'GEOID'.
> Reply example:
> ```python',
> def Load_csv(tract_population_csv_url="https://github.com/gladcolor/LLM-
> Geo/raw/master/overlay_analysis/NC_tract_population.csv"):
> # Description: Load a CSV file from a given URL
> # tract_population_csv_url: Tract population CSV file URL
> tract_population_df = pd.read_csv(tract_population_csv_url)
> return tract_population_df
> ```
>
> Your reply needs to meet these requirements:
>  1. The function description is: Join tract GeoDataFrame with population DataFrame
> 2. The function definition is: join_tract_pop(nc_tract_gdf=nc_tract_gdf, nc_tract_pop_df=nc_tract_pop_df)
> 3. The function return line is: return nc_tract_pop_gdf
> 4. DO NOT change the given variable names and paths.

5. Put your reply into a Python code block(enclosed by ```python and ```), NO explanation or conversation outside the code block.
6. If using GeoPandas to load zipped ESRI files from a URL, load the file directly, DO NOT unzip ESRI files. E.g., gpd.read_file(URL)
7. Generate descriptions for input and output arguments.
8. You need to receive the data from the functions, DO NOT load in the function if other functions have loaded the data and returned it in advance.
9. Note module 'pandas' has no attribute 'StringIO'
10. Use the latest Python module methods.
11. When doing spatial analysis, convert the involved layers into the same map projection.
12. When joining tables, convert the involved columns to string type without leading zeros.
13. When doing spatial joins, remove the duplicates in the results. Or please think about whether it needs to be removed.
14. If using colorbar in GeoPandas maps, set the colorbar's height or length as the same as the map.

The ancestor function code is (need to follow the generated file names and attribute names):

```python
import geopandas as gpd
def load_nc_tract_shp(nc_tract_shp_url='https://github.com/gladcolor/LLM-Geo/raw/master/overlay_analysis/tract_shp_37.zip'):
    """
    Load NC tract boundary shapefile.
    Args:
    nc_tract_shp_url (str): URL of the NC tract shapefile in zip format
    Returns:
    nc_tract_gdf (GeoDataFrame): GeoDataFrame containing NC tracts data
    """
    nc_tract_gdf = gpd.read_file(nc_tract_shp_url)
    return nc_tract_gdf
import pandas as pd
def load_nc_tract_pop_csv(nc_tract_pop_csv_url='https://github.com/gladcolor/LLM-Geo/raw/master/overlay_analysis/NC_tract_population.csv'):
    """
    Load NC tract population CSV file.
    Params:
        nc_tract_pop_csv_url (str): URL of the NC tract population CSV file.
    Returns:
        nc_tract_pop_df (pd.DataFrame): Dataframe containing NC tract population data.
    """
    nc_tract_pop_df = pd.read_csv(nc_tract_pop_csv_url)
    return nc_tract_pop_df
```

The descendant function definitions for the question are (node_name is function name):
 {'node_name': 'calculate_pop_within_tracts', 'description': 'Calculate total population within tracts containing hazardous waste facilities', 'function_definition': 'calculate_pop_within_tracts(haz_waste_gdf=haz_waste_gdf, nc_tract_pop_gdf=nc_tract_pop_gdf)', 'return_line': 'return total_pop_within_tracts'}
{'node_name': 'generate_map', 'description': 'Generate the map showing spatial distribution of population and highlighting borders of tracts with hazardous waste facilities', 'function_definition': 'generate_map(haz_waste_gdf=haz_waste_gdf, nc_tract_pop_gdf=nc_tract_pop_gdf)', 'return_line': 'return population_map'}

**One generated operation function code:**

```python
def join_tract_pop(nc_tract_gdf, nc_tract_pop_df):
    """
    Join tract GeoDataFrame with population DataFrame.
```

```python
    Args:
    nc_tract_gdf (GeoDataFrame): GeoDataFrame containing NC tracts data
    nc_tract_pop_df (pd.DataFrame): Dataframe containing NC tract population data
    Returns:
    nc_tract_pop_gdf (GeoDataFrame): GeoDataFrame containing NC tracts data with population
    """
    # Convert columns to strings without leading zeros
    nc_tract_gdf['Tract'] = nc_tract_gdf['Tract'].astype(str).str.strip().str.replace('^0+', '',
regex=True)
    nc_tract_pop_df['GEOID'] = nc_tract_pop_df['GEOID'].astype(str).str.strip().str.replace('^0+',
'', regex=True)
    # Join DataFrames
    nc_tract_pop_gdf = nc_tract_gdf.merge(nc_tract_pop_df, left_on='Tract', right_on='GEOID',
how='left')
    # Set CRS to match NC tract CRS
    nc_tract_pop_gdf = nc_tract_pop_gdf.to_crs(nc_tract_gdf.crs)
return nc_tract_pop_gdf
```

### 3. Assembly code generation

Prompt (note that this prompt is automatically generated by LLM-Geo based on the question and pre-defined guidance):

Your role: A professional Geo-information scientist and developer good at Python.
Your task is: use the given Python functions, return a complete Python program to solve the question:
   1) Find out the total population that lives within a tract that contains hazardous waste facilities. The study area is North Carolina, US.
   2) Generate a map to show the spatial distribution of population at the tract level and highlight the borders of tracts that have hazardous waste facilities.

Requirement:
   1. You can think step by step.
   2. Each function is one step to solve the question.
   3. The output of the final function is the question to the question.
   4. Put your reply in a code block (enclosed by ```python and ```), NO explanation or conversation outside the code block.
   5. Save final maps, if any.
   6. The program is executable.

Data location:
   1. NC hazardous waste facility ESRI shape file location: https://github.com/gladcolor/LLM-Geo/raw/master/overlay_analysis/Hazardous_Waste_Sites.zip.
   2. NC tract boundary shapefile location: https://github.com/gladcolor/LLM-Geo/raw/master/overlay_analysis/tract_shp_37.zip. The tract id column is 'Tract'.
   3. NC tract population CSV file location: https://github.com/gladcolor/LLM-Geo/raw/master/overlay_analysis/NC_tract_population.csv. The population is stored in 'TotalPopulation' column. The tract ID column is 'GEOID'.

Code:
```python
import geopandas as gpd
def load_haz_waste_shp(haz_waste_shp_url='https://github.com/gladcolor/LLM-Geo/raw/master/overlay_analysis/Hazardous_Waste_Sites.zip'):
    """
    Load hazardous waste facility shapefile
    haz_waste_shp_url: string
```

```python
        URL of the hazardous waste facility shapefile in zipped format
    Returns
    -------
    haz_waste_gdf: geopandas.GeoDataFrame
        GeoDataFrame containing the hazardous waste facility shapefile data
    """
    haz_waste_gdf = gpd.read_file(haz_waste_shp_url)
    return haz_waste_gdf
import geopandas as gpd
def load_nc_tract_shp(nc_tract_shp_url='https://github.com/gladcolor/LLM-
Geo/raw/master/overlay_analysis/tract_shp_37.zip'):
    """
    Load NC tract boundary shapefile.
    Args:
    nc_tract_shp_url (str): URL of the NC tract shapefile in zip format
    Returns:
    nc_tract_gdf (GeoDataFrame): GeoDataFrame containing NC tracts data
    """
    nc_tract_gdf = gpd.read_file(nc_tract_shp_url)
        return nc_tract_gdf
import pandas as pd
def load_nc_tract_pop_csv(nc_tract_pop_csv_url='https://github.com/gladcolor/LLM-
Geo/raw/master/overlay_analysis/NC_tract_population.csv'):
    """
    Load NC tract population CSV file.
    Params:
        nc_tract_pop_csv_url (str): URL of the NC tract population CSV file.
    Returns:
        nc_tract_pop_df (pd.DataFrame): Dataframe containing NC tract population data.
    """
    nc_tract_pop_df = pd.read_csv(nc_tract_pop_csv_url)
    return nc_tract_pop_df
def join_tract_pop(nc_tract_gdf, nc_tract_pop_df):
    """
    Join tract GeoDataFrame with population DataFrame.
    Args:
    nc_tract_gdf (GeoDataFrame): GeoDataFrame containing NC tracts data
    nc_tract_pop_df (pd.DataFrame): Dataframe containing NC tract population data
        Returns:
    nc_tract_pop_gdf (GeoDataFrame): GeoDataFrame containing NC tracts data with population
    """
    # Convert columns to strings without leading zeros
    nc_tract_gdf['Tract'] = nc_tract_gdf['Tract'].astype(str).str.strip().str.replace('^0+',
'', regex=True)
    nc_tract_pop_df['GEOID'] =
nc_tract_pop_df['GEOID'].astype(str).str.strip().str.replace('^0+', '', regex=True)
    # Join DataFrames
    nc_tract_pop_gdf = nc_tract_gdf.merge(nc_tract_pop_df, left_on='Tract', right_on='GEOID',
how='left')
    # Set CRS to match NC tract CRS
    nc_tract_pop_gdf = nc_tract_pop_gdf.to_crs(nc_tract_gdf.crs)
    return nc_tract_pop_gdf
import geopandas as gpd
def calculate_pop_within_tracts(haz_waste_gdf, nc_tract_pop_gdf):
    """
    Calculate total population within tracts containing hazardous waste facilities.
    Args:
    haz_waste_gdf (geopandas.GeoDataFrame): GeoDataFrame containing the hazardous waste
facility shapefile data
    nc_tract_pop_gdf (geopandas.GeoDataFrame): GeoDataFrame containing NC tracts data with
population
    Returns:
    total_pop_within_tracts (int): Total population within tracts containing hazardous waste
facilities
    """
    # Ensure both GeoDataFrames have the same CRS
    haz_waste_gdf = haz_waste_gdf.to_crs(nc_tract_pop_gdf.crs)
    # Perform spatial join between hazardous waste facilities and tract population
GeoDataFrames
    spatial_join_gdf = gpd.sjoin(haz_waste_gdf, nc_tract_pop_gdf, how='inner',
op='intersects')
```

```python
        # Remove duplicates in the resulting GeoDataFrame
        spatial_join_gdf = spatial_join_gdf.drop_duplicates(subset='Tract')
        # Calculate total population within tracts containing hazardous waste facilities
        total_pop_within_tracts = spatial_join_gdf['TotalPopulation'].sum()
        return total_pop_within_tracts
import matplotlib.pyplot as plt
import geopandas as gpd
def generate_map(haz_waste_gdf, nc_tract_pop_gdf):
    """
    Generate the map showing spatial distribution of population and highlighting borders of
tracts with hazardous waste facilities.
    Params:
        haz_waste_gdf (geopandas.GeoDataFrame): GeoDataFrame containing the hazardous waste
facility shapefile data.
        nc_tract_pop_gdf (geopandas.GeoDataFrame): GeoDataFrame containing NC tracts data with
population.
    Returns:
        population_map (matplotlib.pyplot.Figure): The population map.
    """
    # Reproject the hazardous waste GeoDataFrame
    haz_waste_gdf = haz_waste_gdf.to_crs(nc_tract_pop_gdf.crs)
    # Perform a spatial join between hazardous waste facilities and tracts
    hazardous_tracts = gpd.sjoin(nc_tract_pop_gdf, haz_waste_gdf, op='contains',
how='inner').drop_duplicates(subset='Tract')
    # Create the base population map
    fig, ax = plt.subplots(figsize=(10, 10))
    population_map = nc_tract_pop_gdf.plot(column='TotalPopulation', cmap='viridis',
legend=True, linewidth=0.5, edgecolor='k', ax=ax)
    hazardous_tracts.boundary.plot(ax=ax, color='red', linewidth=2)
    # Add a title and set the colorbar's height same as the map
    ax.set_title("Population Distribution and Tracts with Hazardous Waste Facilities",
fontsize=16)
    ax.get_legend().set_bbox_to_anchor((1, 0.5))
    population_map.get_figure().colorbar.set_height(ax.get_position().height)
    return population_map
```

**Generated assembly program:**

```python
# Main program
# Step 1: Load data
haz_waste_gdf = load_haz_waste_shp()
nc_tract_gdf = load_nc_tract_shp()
nc_tract_pop_df = load_nc_tract_pop_csv()
# Step 2: Join NC tract data with population data
nc_tract_pop_gdf = join_tract_pop(nc_tract_gdf, nc_tract_pop_df)
# Step 3: Calculate total population within tracts containing hazardous waste facilities
total_pop_within_tracts = calculate_pop_within_tracts(haz_waste_gdf, nc_tract_pop_gdf)
print("Total population within tracts containing hazardous waste facilities:",
total_pop_within_tracts)
# Step 4: Generate and save map
population_map = generate_map(haz_waste_gdf, nc_tract_pop_gdf)
population_map.get_figure().savefig("population_map.png", dpi=300, bbox_inches="tight")
```

4. **The entire generated executable program for Case Study 3: county level COVID-19 death rate analysis and visualization**

```python
import pandas as pd
def load_covid_data(covid_data_url='https://github.com/nytimes/covid-19-data/raw/master/us-
counties-2020.csv'):
    """
    Description: Load COVID-19 data
    Input: covid_data_url (str): COVID-19 data URL, default is the URL provided
    Output: covid_df (DataFrame): Loaded COVID-19 data as a Pandas DataFrame
    """
    covid_df = pd.read_csv(covid_data_url,dtype={"fips": "str"})
```

```python
    return covid_df
import geopandas as gpd

def load_county_boundary(county_boundary_url='https://github.com/gladcolor/spatial_data/raw/master/contiguous_counties.zip'):
    """
    Load county boundary shapefile from a given URL

    Args:
        county_boundary_url (str): County boundary shapefile URL

    Returns:
        county_boundary_gdf (GeoDataFrame): GeoDataFrame with county boundaries
    """
    county_boundary_gdf = gpd.read_file(county_boundary_url)
    return county_boundary_gdf

import pandas as pd
def load_census_data(census_data_url='https://raw.githubusercontent.com/gladcolor/spatial_data/master/Demography/ACS2020_5year_county.csv'):
    """
    Load Census data from a given URL

    Args:
    census_data_url (str): URL of the Census data

    Returns:
    census_df (DataFrame): Pandas DataFrame containing the Census data
    """
    census_columns = ['FIPS', 'Total Population', 'Total Population: 65 to 74 Years', 'Total Population: 75 to 84 Years', 'Total Population: 85 Years and Over']
    census_df = pd.read_csv(census_data_url, usecols=census_columns, dtype={'FIPS': 'str'})

    return census_df
def calculate_death_rate(covid_df):
    """
    Description: Calculate death rate (death/case) using COVID-19 data
    Input: covid_df (DataFrame): Loaded COVID-19 data as a Pandas DataFrame
    Output: covid_death_rate_df (DataFrame): DataFrame with FIPS and death rate (death/case) information
    """

    # Filter the data to 2020.12.31
    covid_df = covid_df[covid_df["date"] == "2020-12-31"]

    # Calculate the death rate
    covid_df["death_rate"] = covid_df["deaths"] / covid_df["cases"]

    # Keep only FIPS and death rate columns
    covid_death_rate_df = covid_df[["fips", "death_rate"]]

    return covid_death_rate_df
def calculate_senior_rate(census_df):
    """
    Calculate senior resident rate using Census data

    Args:
    census_df (DataFrame): Pandas DataFrame containing the Census data with columns 'FIPS',
    'Total Population', 'Total Population: 65 to 74 Years', 'Total Population: 75 to 84 Years',
    'Total Population: 85 Years and Over'

    Returns:
    census_senior_rate_df (DataFrame): Pandas DataFrame containing the senior resident rate with columns 'FIPS' and 'Senior_Rate'
    """
    census_df['Senior_Population'] = census_df['Total Population: 65 to 74 Years'] + census_df['Total Population: 75 to 84 Years'] + census_df['Total Population: 85 Years and Over']
    census_df['Senior_Rate'] = census_df['Senior_Population'] / census_df['Total Population'] * 100
```

```python
    census_senior_rate_df = census_df[['FIPS', 'Senior_Rate']].copy()

    return census_senior_rate_df

import geopandas as gpd
from tqdm import tqdm

def merge_data(county_boundary_gdf, covid_death_rate_df, census_senior_rate_df):
    """
    Description: Merge data on FIPS: county boundaries, COVID-19 death rate, and Census senior resident rate
    Input:
        county_boundary_gdf (GeoDataFrame): GeoDataFrame with county boundaries
        covid_death_rate_df (DataFrame): DataFrame with FIPS and death rate (death/case) information
        census_senior_rate_df (DataFrame): Pandas DataFrame containing the senior resident rate with columns 'FIPS' and 'Senior_Rate'
    Output:
        merged_data_gdf (GeoDataFrame): GeoDataFrame with merged data
    """

    # Convert 'GEOID' and 'fips' columns to string type
    county_boundary_gdf = county_boundary_gdf.astype({"GEOID": "str"})
    covid_death_rate_df = covid_death_rate_df.astype({"fips": "str"})

    # Merge the three datasets
    merged_data_gdf = county_boundary_gdf.merge(covid_death_rate_df, left_on="GEOID", right_on="fips")
    merged_data_gdf = merged_data_gdf.merge(census_senior_rate_df, left_on="GEOID", right_on="FIPS")

    # Drop duplicates
    merged_data_gdf.drop_duplicates(subset="GEOID", inplace=True)

    return merged_data_gdf

import geopandas as gpd
import matplotlib.pyplot as plt

def create_death_rate_map(merged_data_gdf):
    """
    Description: Plot death rate map using 'quantiles' scheme, 'Conus Albers' projection, and 15x10 inch size
    Input:
        merged_data_gdf (GeoDataFrame): GeoDataFrame with merged data containing county boundaries, COVID-19 death rate, and Census senior resident rate

    Output:
        death_rate_map (matplotlib.figure.Figure): Map figure showing the death rate from the given GeoDataFrame
    """

    # Set map projection to 'Conus Albers'
    merged_data_gdf = merged_data_gdf.to_crs("ESRI:102003")

    # Create plot with 15x10 inch size
    fig, ax = plt.subplots(figsize=(15, 10))

    # Plot death rate using 'quantiles' scheme
    merged_data_gdf.plot(column="death_rate", scheme="quantiles", ax=ax, cmap="viridis", legend=True)

    # Style map
    ax.set_title("COVID-19 Death Rate in US Counties (2020)")
    ax.axis("off")
    fig.subplots_adjust(bottom=0.1, left=0.05, right=0.95, top=0.95, wspace=0.0, hspace=-0.4)

    # Save the figure
    death_rate_map = plt.gcf()

    return death_rate_map
```

```python
import matplotlib.pyplot as plt
import seaborn as sns
from scipy import stats

def create_scatter_plot(merged_data_gdf):
    """
    Description: Plot scatter plot showing correlation and trend line of death rate with senior
    resident rate, including r-square and p-value,  with 50% transparency, red regression line, and
    15x10 inch size

    Input:
        merged_data_gdf (GeoDataFrame): GeoDataFrame with merged data containing death rate and
    senior resident rate

    Output:
        scatter_plot (matplotlib.pyplot.Figure): Scatter plot of death rate vs. senior resident
    rate
    """

    # Extract data for scatter plot
    x = merged_data_gdf['Senior_Rate']
    y = merged_data_gdf['death_rate']

    # Calculate r-square and p-value
    slope, intercept, r_value, p_value, std_err = stats.linregress(x, y)

    # Plot the scatter plot and trend line
    plt.figure(figsize=(15, 10))
    sns.scatterplot(x, y, alpha=0.5)

    plt.plot(x, intercept + slope * x, 'r', label=f'R2={round(r_value ** 2, 3)}; P-value={round(p_value, 3)}')
    plt.legend()

    plt.xlabel("Senior Resident Rate (%)")
    plt.ylabel("COVID-19 Death Rate")
    plt.title("Scatterplot of COVID-19 Death Rate vs. Senior Resident Rate")

    scatter_plot = plt.gcf()

    return scatter_plot

def main():
    # Load data
    covid_df = load_covid_data()
    county_boundary_gdf = load_county_boundary()
    census_df = load_census_data()

    # Calculate death rate and senior rate
    covid_death_rate_df = calculate_death_rate(covid_df)
    census_senior_rate_df = calculate_senior_rate(census_df)

    # Merge data
    merged_data_gdf = merge_data(county_boundary_gdf, covid_death_rate_df, census_senior_rate_df)

    # Create and save death rate map
    death_rate_map = create_death_rate_map(merged_data_gdf)
    death_rate_map.savefig("death_rate_map.png")

    # Create and save scatter plot
    scatter_plot = create_scatter_plot(merged_data_gdf)
    scatter_plot.savefig("scatter_plot.png")

if __name__ == "__main__":
    main()
```